\documentclass{article}

\usepackage{booktabs}
\usepackage{multirow}
\usepackage{graphicx}
\usepackage{amsmath}
\usepackage{amsfonts}

\DeclareMathOperator*{\argmax}{argmax}
\DeclareMathOperator*{\argmin}{argmin}

\usepackage[final]{corl_2019} 

\title{Regularizing Model-Based Planning with Energy-Based Models}

\author{
  Rinu Boney \qquad Juho Kannala \qquad Alexander Ilin\\
  Department of Computer Science\\
  Aalto University, Finland\\
  \texttt{firstname.lastname@aalto.fi} \\
}

\begin{document}
\maketitle


\begin{abstract}
Model-based reinforcement learning could enable sample-efficient learning by quickly acquiring rich knowledge about the world and using it to improve behaviour without additional data. Learned dynamics models can be directly used for planning actions but this has been challenging because of inaccuracies in the learned models. 
In this paper, we focus on planning with learned dynamics models and propose to regularize it using energy estimates of state transitions in the environment.
We visually demonstrate the effectiveness of the proposed method and show that off-policy training of an energy estimator can be effectively used to regularize planning with pre-trained dynamics models.
Further, we demonstrate that the proposed method enables sample-efficient learning to achieve competitive performance in challenging continuous control tasks such as Half-cheetah and Ant in just a few minutes of experience.
\end{abstract}

\keywords{model-based reinforcement learning, energy-based learning}


\section{Introduction}

Reinforcement learning deals with taking sequences of actions in previously unknown environments, to maximize cumulative rewards. Deep reinforcement learning combines the power of deep neural networks for function approximation with reinforcement learning algorithms. Such algorithms have recently been successful in solving several challenging problems \cite{mnih2015human, silver2017mastering, schulman2017proximal}. However, these algorithms are often not sample-efficient, typically requiring an unreasonable amount of interactions in an environment \cite{mnih2015human, schulman2017proximal, duan2016benchmarking}. In real world applications, it is often the case that collecting data is expensive, motivating the need for algorithms that can learn from a minimal number of interactions with the environment.

Effectively using a model of the environment for learning could result in a significant increase of sample-efficiency \cite{chua2018deep, lowrey2018plan}. This is because models introduce an inductive bias that the environment behaves in a forward generative causal direction. Models can be directly used for planning actions that maximize the sum of expected rewards in a finite horizon. Previous works have successfully used fast and accurate simulators of the environment for planning, to solve challenging tasks such as control of a Humanoid \cite{lowrey2018plan, tassa2012synthesis}. However, most real-world problems do not come with a fast and accurate simulator and engineering one from scratch is a laborious task. Even if a simulator exists: 1)~maintaining the simulator to accommodate for changes in the real-world is a lot of manual work, and 2)~matching the true state of the environment and the simulator is not trivial. This motivates the need for \emph{learning} dynamics models of the environment. Dynamics models of the environment can be efficiently learned using supervised learning techniques \cite{nagabandi2018neural} and also efficiently adapted to changes in the environment \cite{clavera2018learning}.

Planning with learned dynamics models is challenging because the planner can exploit the inaccuracies of the model to arrive at actions that are imagined to produce highly over-optimistic rewards (see Figure~\ref{f:traj_opt}). In this paper, we propose to regularize model-based planning with an energy-based model trained on the same transitions as the dynamics model. The planning procedure is augmented with an additive cost of minimizing the energy of the imagined transitions. This penalizes the planner from producing trajectories with transitions that are outside the training data distribution (that is, transitions with high energy estimates). We demonstrate that the proposed method is effective at regularizing model-based planning from exploiting inaccuracies of the model. Previous works have proposed to use ensembles of models to deal with this problem \cite{chua2018deep}. We show that the proposed method can further improve the performance of planning on top of pre-trained ensemble of models. Furthermore, we show that the proposed method enables sample-efficient learning to achieve competitive performance in five popular continuous control tasks.


\section{Model-Based Planning}

In this section, we formalize the problem setting as a Markov Decision Process (MDP). At every discrete time-step $t$, the environment is in state $s_t$, the agent takes an action $a_t$ to receive a scalar reward $r_t = r(s_t, a_t)$ and the environment transitions to the next state $s_{t+1}$, following the dynamics $s_{t+1} = f(s_t, a_t)$. The goal of the agent is to choose actions $a_t$ so as to maximize the sum of expected future rewards (called return), $G = \mathbb{E} \left[ \sum_{t=0}^\infty r(s_t, a_t) \right]$.

In this paper, we focus on finite-horizon planning with learned dynamics models $\hat{f}$. Also, we assume that the reward function $r(s_t, a_t)$ is known and that the state $s_t$ is fully observed. Forward dynamics models of the environment can be learned using supervised learning techniques to predict the next state $s_{t+1}$ from the current state $s_t$ and action $a_t$:
\[ s_{t+1} = \hat{f}(s_t, a_t) \,. \]

\begin{figure}[t]
\begin{center}
\includegraphics[width=0.5\textwidth, trim=180 120 180 120, clip]{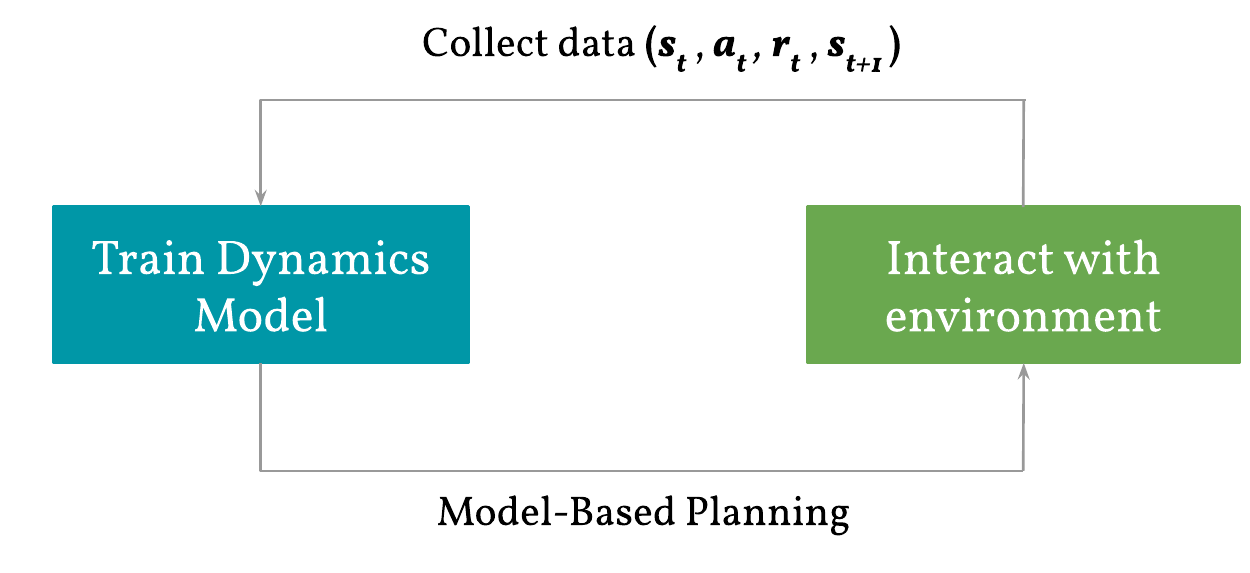}
\end{center}
\caption{Overview of training loop. We initially perform random exploration for one or more episodes to train the dynamics model and then interact with the environment by planning using the learned dynamics model. At the end of each episode, we re-train the dynamics model on the past experience.}
\label{f:training-loop}
\end{figure}

At time-step $t$, the agent can plan a sequence of actions $\{a_t, \ldots, a_{t+H}\}$ by unrolling the learned dynamics model to maximize the sum of rewards:
\begin{align}
a^*_t, \ldots, a^*_{t+H} = \argmax_{a_t, \ldots, a_{t+H}} \mathbb{E} \left[ \sum_{\tau=t}^{t+H} r(s_\tau, a_\tau) \right] 
\,,
\label{eq:obj-true}
\end{align}
such that $s_{\tau+1} = f(s_\tau, a_\tau)$. Since we do not have access to the true dynamics $f$, we plan using the following objective as a proxy to the true objective:
\begin{align}
a^*_t, \ldots, a^*_{t+H} = \argmax_{a_t, \ldots, a_{t+H}} \sum_{\tau=t}^{t+H} r(s_\tau, a_\tau) 
\,,
\label{eq:obj-proxy}
\end{align}
such that $s_{\tau+1} = \hat{f}(s_\tau, a_\tau)$. We use model-predictive control (MPC) \cite{garcia1989model, nagabandi2018neural} to adapt our plans to new states, that is we apply just the first action from the optimized sequence and re-plan at next step. This reduces the effect of error accumulation due to multi-step predictions using the model. 

We initially train the dynamics model using data collected by executing a random policy for one or more episodes. More episodes of initial exploration reduces the effect of the initial model weights. The initial data can also be collected by an existing policy, for example human demonstrations or human-engineered controllers. After training the dynamics models, we interact with the environment using model-based planning (Equation~\ref{eq:obj-proxy}). The $(s_t, a_t, s_{t+1})$ transition observed at each interaction is stored in a replay buffer along with the initial data and the model is re-trained on the replay buffer at the end of each episode. We iterate this alternating process of training the dynamics model and model-based planning. An overview of the training loop is illustrated in Figure~\ref{f:training-loop}.

\subsection{Regularizing Model-Based Planning}

\begin{figure}[t]
\begin{center}
\includegraphics[width=\textwidth, trim=20 80 20 100, clip]{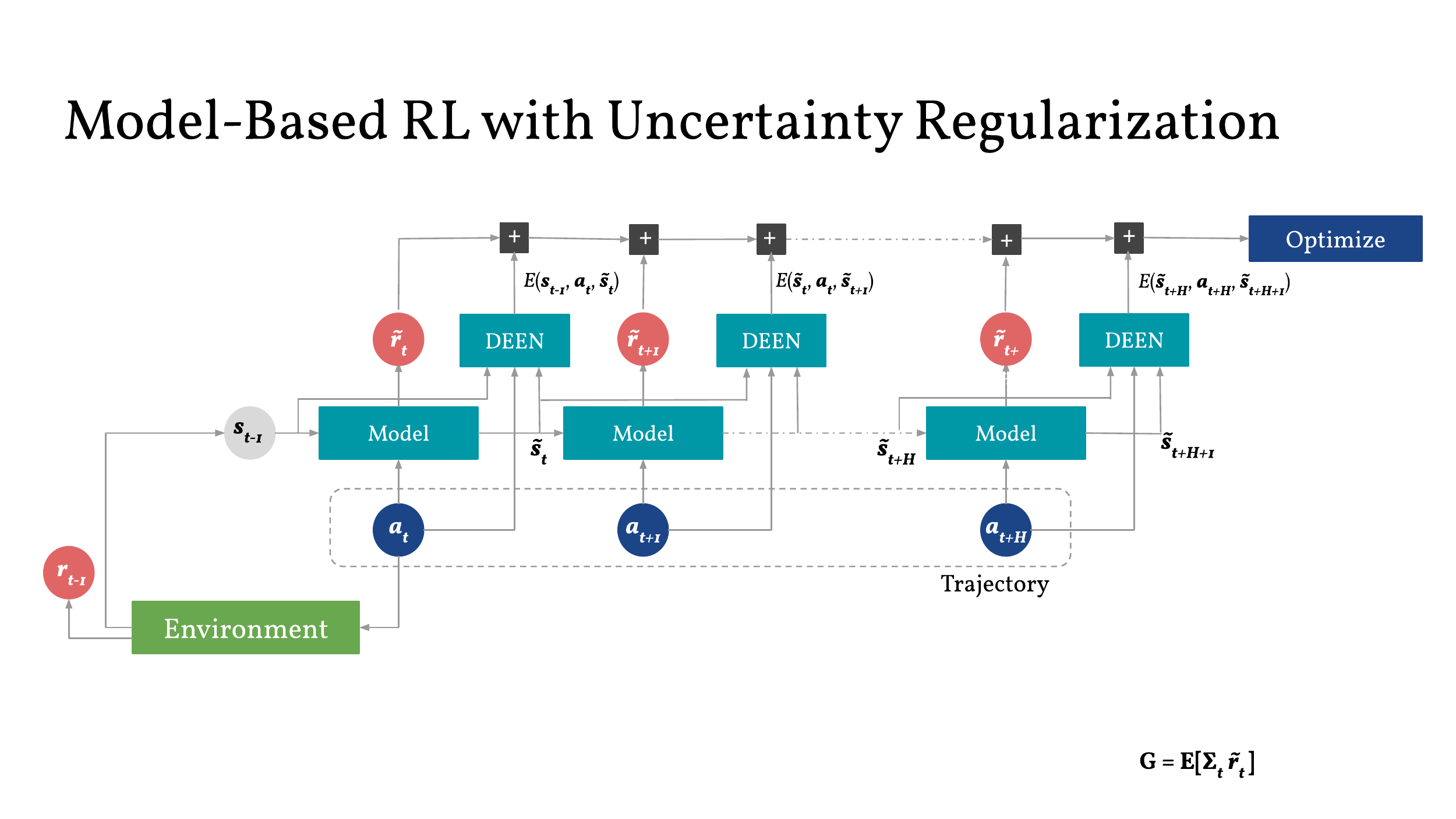}
\end{center}
\caption{Computational graph of regularizing model-based planning with energy-based models. At each timestep $t-1$, the environment is in a state $s_{t-1}$ and we initially consider an action trajectory $\{a_t, a_{t+1}, \ldots, a_{t+H}\}$ that we could apply for the next $H$ timesteps. The dynamics model is used to predict the future state transitions $\{\tilde{s}_t, \tilde{s}_{t+1}, \ldots, \tilde{s}_{t+H}\}$ at each timestep for the considered action trajectory. The rewards at each time-step is computed directly from the state-action pairs, resulting in a prediction of the finite-horizon cumulative reward. An additive regularization term is augmented to the planning objective by computing the energy of each $(\tilde{s}_\tau, a_\tau, \tilde{s}_{\tau+1})$ transition. The action trajectory $\{a_t, a_{t+1}, \ldots, a_{t+H}\}$ can be optimized to maximize this regularized planning objective.}
\label{f:comp-graph}
\end{figure}

Directly optimizing the objective in Equation~\ref{eq:obj-proxy} is challenging because $\hat{f}$ is only an approximation of the true dynamics $f$. Deep neural networks are commonly used as function approximators to learn $\hat{f}$ and they are amenable to erroneous predictions for samples outside the data distribution. 
An effective optimizer can be easily deceived by these erroneous predictions, converging to action trajectories that are imagined to produce very high rewards but is not the case in reality. This problem can be alleviated by penalizing the optimizer from considering trajectories that are outside the training distribution of the learned dynamics model $\hat{f}$. This can be achieved by augmenting the planning objective with an additive term to maximize the probability of imagined transitions:
\begin{align}
a^*_t, \ldots, a^*_{t+H} = \argmax_{a_t, \ldots, a_{t+H}} \sum_{\tau=t}^{t+H} r(s_\tau, a_\tau) + 
\alpha \log p(s_t, a_t, s_{t+1}, \ldots, s_{t+H}, a_{t+H}, s_{t+H+1})
\,.
\nonumber
\end{align}
where the scalar $\alpha$ modulates the weight between both costs. We approximate the joint probability of the whole trajectory as a sum of joint probabilities of each transition in the trajectory:
\begin{align}
a^*_t, \ldots, a^*_{t+H} = \argmax_{a_t, \ldots, a_{t+H}} \sum_{\tau=t}^{t+H} \left[ r(s_\tau, a_\tau) +
\alpha \log p(s_\tau, a_\tau, s_{\tau+1}) \right]
\,.
\label{eq:obj-main}
\end{align}

Assume that we want to learn the probability density function using a parameterized model $p(x_\tau; \theta)$, where $x_\tau = [s_\tau, a_\tau, s_{\tau+1}]$. The energy function $E(x_\tau; \theta)$ is the unnormalized log-density, that is the probability density function $p(x_\tau)$ is defined as:
\begin{align}
p(x_\tau; \theta) = \frac{1}{Z(\theta)} \exp{(-E(x_\tau;\theta))}
\,,
\nonumber
\end{align}
where $Z(\theta) = \int{\exp{(-E(x'_\tau;\theta))}} dx'_\tau$ is the partition function which normalizes the probability density. Computing the partition function is generally intractable in practice and is not important for regularization in Equation~\ref{eq:obj-main} since it does not depend on $x_\tau$.
We can instead learn and use the energy function for regularizing model-based planning:
\begin{align}
a^*_t, \ldots, a^*_{t+H} = \argmax_{a_t, \ldots, a_{t+H}} \sum_{\tau=t}^{t+H} \left[ r(s_\tau, a_\tau) -
\alpha E(s_\tau, a_\tau, s_{\tau+1}) \right]
\,.
\label{eq:obj-energy}
\end{align}


\section{Energy-Based Models}
\label{sec:energy}

In principle, any energy-based model \cite{lecun2006tutorial} can be used to estimate $E(s_\tau, a_\tau, s_{\tau+1})$ in Equation~\ref{eq:obj-energy}. In this paper, we use the recently introduced Deep Energy Estimator Networks (DEEN) \cite{saremi2018deep, saremi2019neural} for energy estimation. In this section, we introduce deep energy estimator networks and further contrast it against direct score function estimation using a denoising autoencoder. We show that deep energy estimator networks offers a principled and scalable way to estimate both the energy and score functions, making it a good choice for regularization in both gradient-free and gradient-based planning.

Consider a random variable $Y$ that is a noisy observation of another unknown random variable $X$ with density function $p(x)$. 
\citet{robbins1956empirical} derived the least squares estimators of variable $X$ from observed value $y$ of variable $Y$ for Poisson, geometric and binomial noise distributions and \citet{miyasawa1961empirical} later extended the work to derive the estimator for univariate Gaussian noise. \citet{raphan2011least} generalized these results into a unified framework and derived least square estimators for more distributions including multivariate Gaussian distribution. The empirical Bayes least squares estimator or the optimal denoising function $g(y)$ for zero-mean multivariate Gaussian noise is given by:
\begin{align}
g(y) = y + \sigma^2 \nabla_y \log p(y)
\,,
\label{eq:opt-denoising}
\end{align}
where $y \sim x + N(0, \sigma^2I_d)$. Assume that we have access to samples $x_i \in X$. Then, we can corrupt the samples using zero-mean Gaussian noise to obtain samples $y_i \in Y$. We can train a feedforward neural network $\hat{g}$ to denoise each sample $y_i$ to predict $x_i$. Such a function $\hat{g}$ can be implemented with a denoising autoencoder (DAE)
and based on Equation~\ref{eq:opt-denoising} we can use it to approximate the score function $\nabla_y \log p(y)$ of the corrupted distribution as follows:
\begin{align}
\nabla_y \log p(y) \propto \hat{g}(y) - y
\,.
\label{eq:dae}
\end{align}

\begin{figure}[t]
\begin{center}
\includegraphics[width=\textwidth, trim=10 10 10 10, clip]{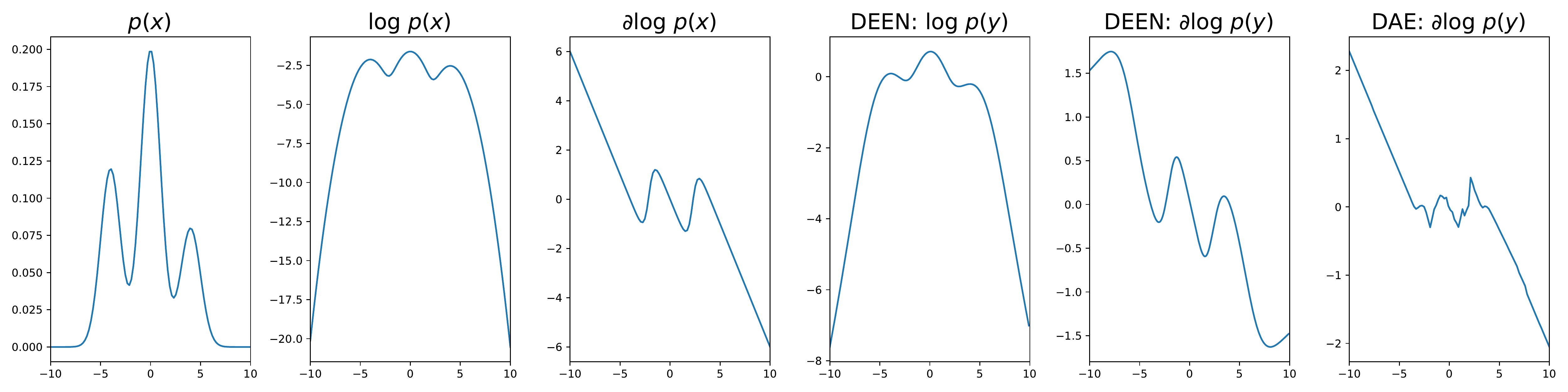}
\end{center}
\caption{Comparison of score function estimation using DEEN vs DAE. We consider a simple mixture of 1d Gaussian distributions and generate 1000 samples from it. The ground truth probability density function $p(x)$, $\log\ p(x)$ and score function $\frac{\partial\log\ p(x)}{\partial x}$ are shown in the first three figures. We corrupt the training data using additive Gaussian noise with a scale of 0.5. We train a DEEN and a DAE on this training data and the energy and score function estimates of the corrupted distribution $p(y)$ are shown in the last three figures. DEEN provides good smooth estimates of the energy and score function. DAE also provides a reasonable estimate of the score function.}
\label{f:deen-dae}
\end{figure}

\citet{boney2019regularizing} proposed to use this approximation to directly estimate the gradient of $\log p(s_\tau, a_\tau, s_{\tau+1})$ in Equation~\ref{eq:obj-main}. This can be used in gradient-based planning by using the penalty term $\|\hat{g}(x) - x\|^2$ instead of $\log p(s_\tau, a_\tau, s_{\tau+1})$ in Equation~\ref{eq:obj-main} and stopping the gradient propagation through the denoising autoencoder $g$. 

Gradient-free planning with the regularized objective in Equation~\ref{eq:obj-energy} requires explicit energy estimates. \citet{boney2019regularizing} used DAE regularization for gradient-free planning, which is not accurate as the denoising error does not correspond to energy estimation. Gradient-free planning offers some attractive properties: 1)~It is very easy to parallelize, 2)~There is no need to backpropogate gradients, 3)~Gradient-based planning involves backpropagating through very deep networks (where the same network is repeatedly applied for a finite horizon), where the problem of exploding and vanishing gradients may arise, and 4)~The learned dynamics model can become chaotic leading to high variance in the backpropagated gradients \cite{parmas2018pipps}. In this paper, we propose to use differentiable energy-based models to obtain explicit energy estimates, from which the score function can be computed (if needed).

\citet{saremi2018deep} proposed to explicitly parameterize the energy function $E(y; \theta)$ with a neural network
and to compute the derivative of the energy by backpropagating through the network. Such a network can be trained by minimizing the following objective based on the relation in Equation~\ref{eq:opt-denoising}:
\begin{align}
\argmin_\theta \sum_{x_i \in X, y_i \in Y} \left\| x_i - y_i + \sigma^2 \frac{\partial E(y=y_i; \theta)}{\partial y} \right\|^2
\label{eq:deen}
\end{align}

The energy function network $E(y; \theta)$ is called deep energy estimator network (DEEN).
Note that minimizing this objective involves double backpropagation at each step. Optimizing the objective in Equation~\ref{eq:deen} ensures that the score function $\partial E(y; \theta) / \partial y$ satisfies the relation in Equation~\ref{eq:opt-denoising}. This leads the energy network $E$ to explicitly learn the energy function of the corrupted distribution such that the gradient of the network also corresponds to the score function of the corrupted distribution. In this paper, we propose to use the energy network $E(y; \theta)$ to learn the energy function $E(s_\tau, a_\tau, s_{\tau+1})$ and use it for regularization in the planning objective (Equation~\ref{eq:obj-energy}).  A computational graph of regularizing model-based planning with energy estimation using a DEEN network is illustrated in Figure~\ref{f:comp-graph}.

It is to be noted that both the denoising autocoder and the DEEN methods approximate the score function of the corrupted distribution $p(y)$ instead of the true data distribution $p(x)$. This can potentially behave better in practice since $p(y)$ can be seen as a Parzen window estimate of $p(x)$ with variance $\sigma^2$ as the smoothing parameter \cite{saremi2019neural, vincent2011connection}.

In Section~\ref{sec:exp}, we show that energy estimation with DEEN is more effective than direct score function estimation using DAE for regularizing model-based planning. Deep energy estimator networks have been shown to be robust for score function estimation since the score function is computed from explicit energy estimates \cite{saremi2018deep}. We compare score function estimation using denoising autoencoders and deep energy estimator networks on a toy example in Figure~\ref{f:deen-dae}. Previous works \cite{saremi2019neural, alain2014regularized} have also observed that directly estimating the score function is not robust in practice.


\section{Experiments}
\label{sec:exp}

We compare the proposed energy-based regularization method to probabilistic ensembles with trajectory sampling (PETS) \cite{chua2018deep} and DAE regularization \cite{boney2019regularizing}. PETS is a state-of-the-art model-based algorithm which involves learning an ensemble of probabilistic dynamics models. 
We perform the comparison in five popular continuous control benchmarks from \cite{brockman2016openai}: Cartpole, Reacher, Pusher, Half-cheetah and Ant. We use cross-entropy method (CEM) \cite{botev2013cross} as the optimizer in all our experiments since it is computationally significantly faster than the Adam optimizer used in \cite{boney2019regularizing} and also was able to achieve competitive or even better results in these benchmarks. In Section~\ref{sec:exp-pre}, we test energy-based regularization method on top of the pre-trained PETS models 
to show that energy-based regularization further improves planning. In Section~\ref{sec:exp-scratch}, we 
show that enery-based regularization enables sample-efficient learning to solve all tasks from just a handful of trials.

\subsection{Experiments on Pre-Trained Dynamics Models}
\label{sec:exp-pre}

We test the proposed regularization on top of the state-of-the-art model-based RL algorithm: PETS. We trained an ensemble of probabilistic dynamics models using the code provided by the authors of \cite{chua2018deep}. The results are shown in Table~\ref{t:pretrained-results}. We trained PETS on the Half-cheetah benchmark for 300 episodes and perform closed-loop planning by augmenting the planning objective with an additive term consisting of the energy estimates (Equation~\ref{eq:obj-energy}). Both DAE and DEEN regularization are able to improve upon PETS, with DEEN regularization performing the best. Similar to \cite{boney2019regularizing}, we did not observe any improvements on tasks with low-dimensional action spaces: Cartpole, Reacher and Pusher.


\begin{table}[h]
\caption{Comparison of planning using pre-trained PETS models with different optimizers}
\label{t:pretrained-results}
\centering
\begin{tabular}{llllll}  
\toprule
Optimizer & CEM & CEM + DAE & Adam + DAE & CEM + DEEN \\
\midrule
Return & $10955 \pm 2865$ & $12967 \pm 3216$ & $12796 \pm 2716$ & $\mathbf{13052 \pm 2814}$ \\
\bottomrule
\end{tabular}
\end{table}


\subsection{Experiments on Learning from Scratch}
\label{sec:exp-scratch}

\begin{figure}[t]
\begin{center}
\includegraphics[width=\textwidth]{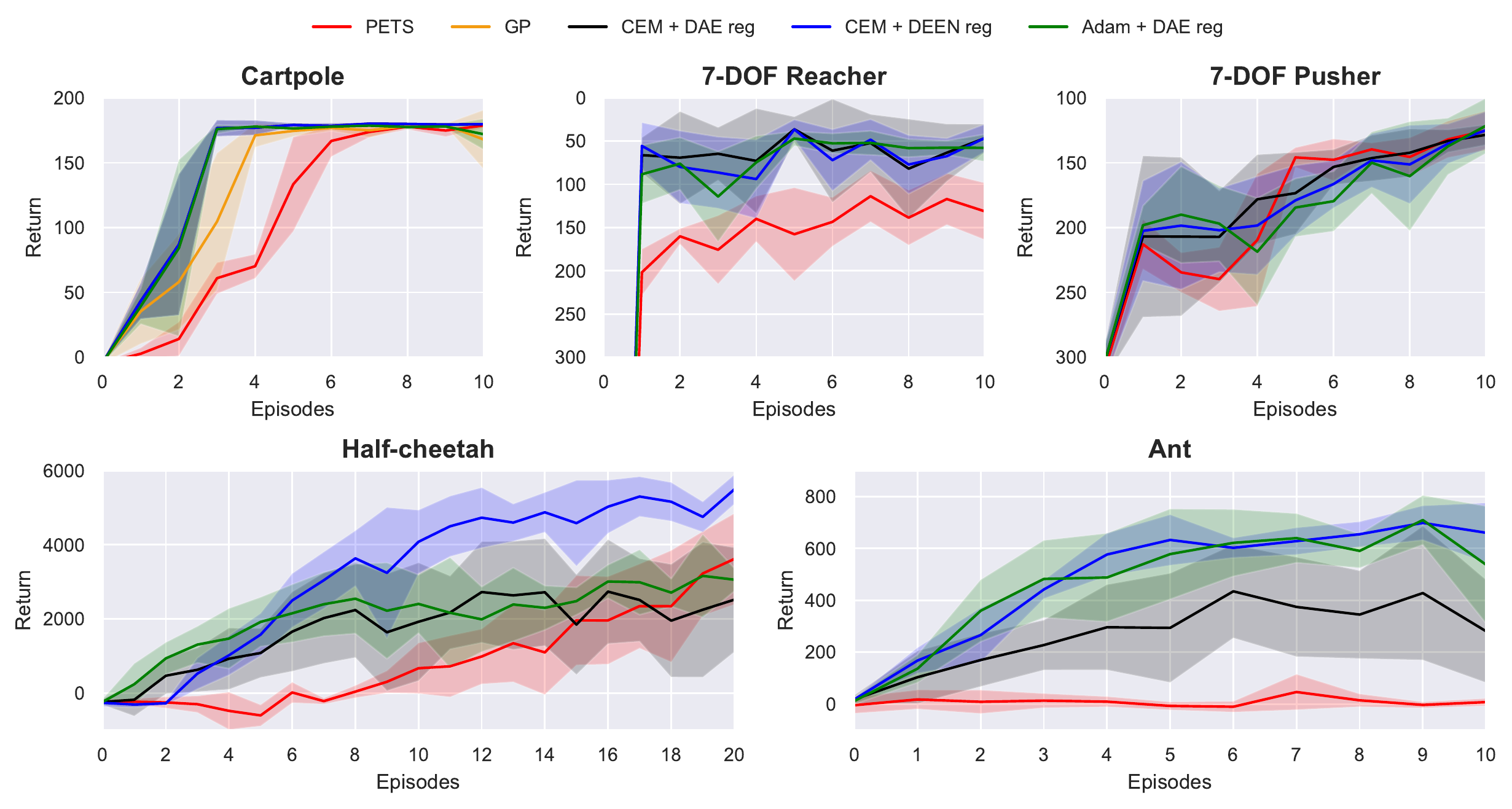}
\end{center}
\caption{Results of our experiments on learning from scratch on five continuous control benchmarks: Cartpole, Reacher, Pusher, Half-cheetah and Ant. We compare to PETS \cite{chua2018deep} and DAE regularization \cite{boney2019regularizing} using the return (cumulative reward) obtained in each episode. 
PETS is a state-of-the-art model-based RL algorithm and DAE regularization has been shown to be effective in sample-efficient learning in these tasks. We also compare against planning with Gaussian Processes (GP) based dynamics model in the Cartpole task. We show the mean and standard deviation of each setting averaged across 5 seeds.}
\label{f:scratch-results}
\end{figure}

In this section, we test the effectiveness of regularization using energy-based models for learning from scratch. We perform one or more episodes of random exploration, train a dynamics model and the regularizer on the transitions and further interact with the environment using MPC by planning at each time-step using the regularized objective in Equation~\ref{eq:obj-energy}. We maintain a replay buffer of all past transitions and at the end of each episode the dynamics model and regularizer are re-trained on the whole replay buffer. In these experiments, we use a single feedforward network as the dynamics model (as opposed to an ensemble of models in \cite{chua2018deep}) to test the efficacy of the proposed method in a simple setting. The results are shown in Figure~\ref{f:scratch-results}. In low-dimensional environments such as Cartpole, Reacher and Pusher, CEM optimization with DEEN regularization is comparable or better than the other methods. In Half-cheetah, CEM optimization with DEEN regularization clearly performs the best, obtaining good asymptotic performance in just 20,000 timesteps (corresponding to 16.6 minutes of experience). In Ant, CEM optimization with DEEN regularization also performs the best, learning to walk reasonably in just 4,000 timesteps (corresponding to 3.3 minutes of experience)\footnote{Videos of the training progress are available at \href{https://sites.google.com/view/regularizing-mbrl}{https://sites.google.com/view/regularizing-mbrl}.}. It can also be observed that CEM optimization with DEEN regularization performs competitively or better than Adam optimization with DAE regularization, which requires much more computation.
In the Half-cheetah benchmark, state-of-the-art model-free methods \cite{haarnoja2018soft, fujimoto2018addressing} and PETS obtains better asymptotic performance during later stages of learning. 
We postulate that this is due to lack of a proper exploration mechanism in the proposed approach. However, DEEN regularization enables excellent performance during the early stages of training and also shows consistent improvements after each episode. This is very important for practical applications in the real-world, where the agent is expected to perform sensible actions from the very beginning. The proposed method enables efficient exploitation of the learned dynamics model and combining it with an explicit exploration mechanism would facilitate controlled exploration and exploitation. We leave complementation of the proposed method with an effective exploration strategy to future research. For example, energy estimates of the transitions could be used as bonuses for curiosity-driven exploration \cite{pathak2017curiosity} to visit novel states and try novel actions.

\begin{figure}[t]
\begin{center}
\includegraphics[width=.8\textwidth]{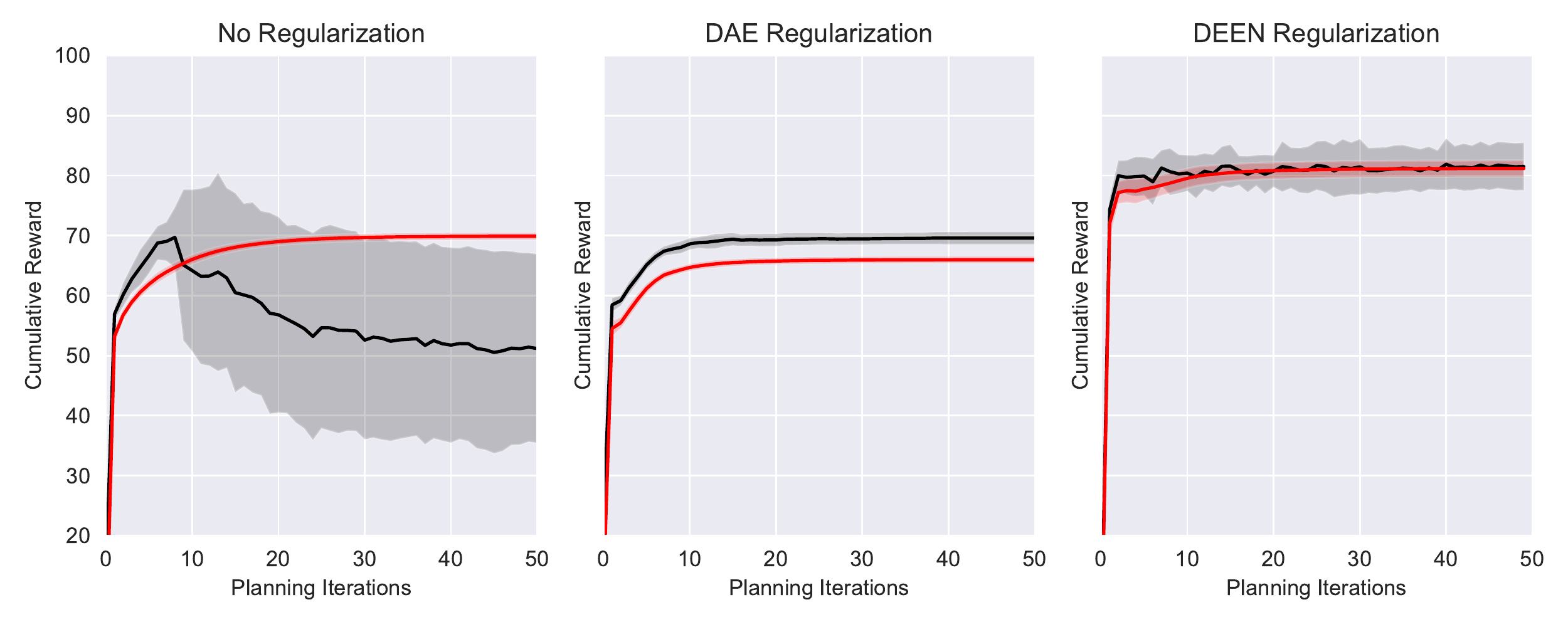}
\end{center}
\caption{Visualization of trajectory optimization after 5 episodes of training in the Half-cheetah environment. We perform actions in the environment for 50 timesteps using MPC and then initialize the CEM optimization with the a random population of trajectories and optimize the planning objective in three different settings: 1) without any regularization, 2) with DAE regularization \cite{boney2019regularizing}, and 3) with DEEN regularization. Here, the red lines denote the rewards predicted by the model (imagination) and the black lines denote the true rewards obtained when applying the sequence of optimized actions (reality). It is noticeable that planning without any regularization exploits inaccuracies of the dynamics model but DAE and DEEN regularization is able to prevent this.
}
\label{f:traj_opt}
\end{figure}

We visually demonstrate the effectiveness of DEEN regularization in Figure~\ref{f:traj_opt}. To compare with \cite{boney2019regularizing}, we visualize trajectory optimization on the Half-cheetah task using dynamics models obtained after 5 episodes of training. We perform actions in the environment for 50 timesteps using model-based planning to arrive at a stable state and then perform trajectory optimization using a randomly initialized population of action trajectories. Without any regularization, the planning procedure leads to trajectories that are predicted to produce high rewards but is not the case in reality. It can be observed that while DAE regularization is also effective, DEEN regularization is clearly better, being able to successfully prevent the reality and imagination from diverging and also leading to trajectories with a better outcome.

\subsection{Model Architecture and Hyperparameters}

The important hyperparameters used in our experiments are reported in Table~\ref{t:hyperparameters}.
Following \cite{chua2018deep, boney2019regularizing}, we used a Bayesian neural network (BNN) with mean and variance predictions as our dynamics model. Although the predicted variance is not used for planning, it was found to have a regularizing effect during training \cite{chua2018deep}. We used a vanilla feedforward network to model the energy function.
The energy estimation network is trained by corrupting the transitions in the replay buffer using additive Gaussian noise and minimizing the objective in Equation~\ref{eq:deen}. We found the noise scale $\sigma$, cost multiplier $\alpha$ and number of training epochs to be the most sensitive hyperparameters. 
To prevent overfitting to the replay buffer, we explored simple strategies like increasing the number of initial episodes with random exploration and decaying the number of training epochs after each episode. In Half-cheetah, we perform random exploration for the first 3 episodes and decay the number of training epochs to 8 after 10 episodes. In Ant, we decay the number of training epochs of the dynamics model and the DEEN network by factors of 0.6 and 0.7 respectively.

\begin{table}[h]
\caption{Important hyperparameters used in our experiments}
\label{t:hyperparameters}
\centering
\begin{tabular}{llccccc}
\toprule
& Hyperparameter & Cartpole & Reacher & Pusher & Half-cheetah & Ant \\
\midrule
\multirow{4}{*}{Model} & Hidden layers & 3 & 3 & 3 & 4 & 4 \\
& Hidden size & 200 & 200 & 200 & 200 & 400 \\
& Epochs & 500 & 500 & 100 & 300 & 600 \\
& Batch Size & 32 & 32 & 32 & 128 & 400 \\
\midrule
\multirow{6}{*}{DEEN} & Hidden layers & 3 & 3 & 3 & 5 & 3 \\
& Hidden size & 200 & 200 & 200 & 500 & 300 \\
& Epochs & 500 & 500 & 100 & 100 & 800 \\
& Batch Size & 32 & 32 & 32 & 32 & 64 \\
& Noise scale $\sigma$ & 0.1 & 0.1 & 0.1 & 0.37 & 0.9 \\
& Cost multiplier $\alpha$ & 0.001 & 0.001 & 0.01 & 0.05 & 0.035 \\
\midrule
CEM & Iterations & 5 & 5 & 5 & 5 & 7 \\
\bottomrule
\end{tabular}
\end{table}


\section{Related Work}

This paper is inspired by the recent works on planning with learned dynamics models. PILCO \cite{deisenroth2011pilco} is a well-known method for sample-efficient learning in low-dimensional control problems. However, it has been difficult to scale such methods to high-dimensional problems. \citet{nagabandi2018neural} used neural networks as dynamics models for planning to demonstrate sample-efficient learning on challenging continuous control problems and further fine-tuned the learned policy using model-free methods for good asymptotic performance. \citet{mishra2017prediction} introduced deep generative models based on convolutional autoregressive models and variational autoencoders \cite{kingma2013auto} and used it for gradient-based trajectory optimization and policy optimization. \citet{chua2018deep} used ensembling of probabilistic dynamics models and trajectory sampling to deal with inaccuracies of learned dynamics models and demonstrated sample-efficient learning to achieve high asymptotic performance on challenging continuous control problems. While the use of ensembling in \cite{chua2018deep} allows for a better dynamics model, energy-based regularization also bounds the transitions of the agent to be similar to it's previous experience, which might be important for safety-critical applications. While KL divergence penalty between action distribution has been used in previous works \cite{levine2013guided, kumar2016optimal}, energy-based regularization also bounds the familiarity of states. \citet{boney2019regularizing} proposed to use denoising autoencoders to prevent trajectory optimization from exploiting modelling inaccuracies and demonstrated sample-efficient learning from very few episodes, using gradient-based trajectory optimization. \citet{hafner2018learning} introduced a novel latent dynamics model for planning in high-dimensional observation spaces, such as images. Recent works on model-based RL have also explored Dyna-style \cite{sutton1990integrated} architectures where learned dynamics models are used to generate data to train model-free methods \cite{kurutach2018modelensemble, clavera2018model, ha2018recurrent}.

Energy-based models can be directly used for planning by sampling future trajectories from the generative model. \citet{du2019implicit} showed that energy-based models can be used to sample diverse predictions of future trajectories and it was later extended in \cite{du2019model} for model-based planning. DEEN \cite{saremi2018deep} could also be directly used for planning by sampling future  trajectories using the novel walk-jump sampling algorithm introduced in \cite{saremi2019neural}. However, sampling from such models at each timestep for planning is expensive and we instead use a separate forward dynamics model for directly predicting the future trajectories but only use the energy-based model for regularization. This can be seen as an ensemble of two different kinds of models.


\section{Conclusion}

Planning with learned dynamics models is challenging because planning can exploit the inaccuracies in the model to produce over-optimistic trajectories. In this paper, we propose to regularize planning using energy estimates of state transitions in the environment. We use a recently proposed energy estimation method called DEEN for this purpose. We demonstrated that an energy estimation network can be trained on the past experience of pre-trained dynamics models to further improve planning. We also demonstrated that the energy regularization enables sample-efficient learning on challenging tasks such as Half-cheetah and Ant, in just a few minutes of interaction.

One of the limitations of the proposed and related model-based planning algorithms is the additional hyperparameter tuning required for learning dynamics models. AutoML algorithms can be potentially used to automate the training of effective dynamics model by splitting the replay buffer into a training set and validation set and optimizing the prediction performance on the validation set. This could enable automatic architecture and hyperparameter tuning of dynamics models using more computational resources, without any additional data or human supervision. This would be an interesting line of future work.



\clearpage

\acknowledgments{We would like to thank Saeed Saremi for valuable discussions about his work on deep energy estimator networks and neural empirical Bayes.}


\bibliography{example}  

\end{document}